\documentclass[runningheads]{llncs}
\usepackage[T1]{fontenc}
\usepackage{graphicx}
\usepackage{comment}
\usepackage{subcaption}
\usepackage[noadjust]{cite} 
\usepackage{xcolor} 
\usepackage{amsmath,amssymb} 
\usepackage{booktabs}
\usepackage{siunitx}
\usepackage{threeparttable}
\usepackage{algorithm}
\usepackage{algpseudocode}

\newcommand{\luki}[1]{\textcolor{red}{[Luki: #1]}}

\begin{document}
\title{Adaptive Repulsive Pheromone Clustering for Foraging Robot Swarms}
\titlerunning{Adaptive Repulsive Pheromone Clustering}
%
\author{Carlos Pena-Caballero\orcidID{0009-0003-6576-3033}\and Constantine Tarawneh\orcidID{0000-0002-4074-5627}\and
 Qi Lu\orcidID{0000-0002-8425-4686}}
\authorrunning{C. Pena-Caballero and Q. Lu}
%
\institute{The University of Texas Rio Grande Valley, Edinburg, TX, USA
\email{\{carlos.penacaballero01,constantine.tarawneh,qi.lu\}@utrgv.edu}}
\maketitle              
\begin{abstract}

The Central Place Foraging Algorithm (CPFA) combines site fidelity, pheromone-guided navigation, and uninformed random search to enable decentralized resource collection in robot swarms. However, CPFA often revisits previously explored regions while leaving other areas insufficiently searched, reducing efficiency as resources become scarce. In this paper, we propose Adaptive Repulsive Pheromone Clustering (ARPC), a bio-inspired method in which robots deposit repulsive pheromone waypoints to mark previously explored locations. These waypoints are clustered around the nest to estimate low-value search regions, allowing robots to be redirected toward likely unvisited areas. By integrating the exploitation of known resources with systematic avoidance of redundant exploration, ARPC improves search diversity and resource discovery efficiency. Extensive simulations in ARGoS across varying arena sizes, resource densities, and clustered, random, and power-law spatial distributions demonstrate that ARPC consistently outperforms CPFA and the Grid-Based CPFA (GPFA). In particular, ARPC yields significant gains during both early discovery (10\%) and late-stage (up to 60\%) collection, where conventional methods typically degrade. These results indicate that ARPC provides a scalable and robust strategy for large-scale heterogeneous swarm foraging environments.

\end{abstract}
%
%
\section{Introduction}

Swarm robotics is a research field that explores the coordination of multiple simple and autonomous robots to perform complex tasks in a decentralized way \cite{navarro2012introduction}. Contemporary research explores a range of collective behaviors, including self-organization \cite{self-org2}, \cite{self-org1}, task allocation \cite{acbba2020}, aggregation \cite{cue-based1}, \cite{cue-based2}, object sorting \cite{sorting0}, \cite{sorting1}, foraging \cite{BeyondPherom2015, ForagingDRL2020, QualitySensitiveForaging2018, arturo2025}.

The Central Place Foraging Algorithm (CPFA) \cite{BeyondPherom2015} is a bio-inspired swarm robotics method that captures the group foraging dynamics of desert harvester ants (Pogonomyrmex barbatus)\cite{ants}. It integrates site fidelity, pheromone-based guidance, and uninformed random search to achieve decentralized resource gathering by a swarm of robots. Despite its effectiveness, the standard CPFA exhibits spatial blindness: robots tend to repeatedly revisit familiar regions while neglecting other areas that may contain abundant resources. In \cite{CompleteCollection2015}, experimental results indicate that CPFA spends 63\% and 75\% of its total run time collecting the final 12\% of resources. 

Being able to quickly and cheaply recall regions of interest is crucial for speeding up resource collection. However, in real-world settings, this information is usually unavailable, so we need a mechanism to quickly exclude areas that do not warrant further exploration. In this paper, we propose the Adaptive Repulsive Pheromone Cluster algorithm (ARPC), an algorithm that can rapidly identify and eliminate locations that no longer justify exploration, thereby reducing the time needed to gather the remaining 12\% of resources. In our method, robots lay down repulsive pheromones along their paths to mark and subsequently avoid already explored regions. They then report these repulsive pheromone positions back to the nest, where we use them to build a clustered map of zones that no longer require exploration. We then take the complement of this map to identify never-visited locations and dispatch robots specifically to those areas. The design of this algorithm will be detailed in Section \ref{methods}. To validate our approach, we will perform three experiments to evaluate its performance across different arena sizes, resource distributions, and total quantities of available resources.

ARPC targets a core practical challenge: enabling large teams to coordinate exploration with a single collaborative generated map of “negative information” (repulsive pheromones) that can be aggregated intermittently at the central depot, the approach is compatible with realistic swarm constraints such as limited computation, intermittent communication, and noisy sensing.


The rest of the paper is organized as follows: in Section \ref{related}, we explore related works trying to improve on the CPFA algorithm; in Section \ref{cpfa} Section we introduce the original CPFA algorithm; \ref{methods} discusses the new algorithm ARPC; Section~\ref{experiments} explains the experimental setup for ARPC; and in Section~\ref{conclusion}, we give our summary of our contributions and future work.



%
\section{Related Work}
\label{related}

Numerous projects have sought to enhance the foraging efficiency of decentralized robotic systems. The CPFA algorithm introduced site fidelity and pheromone trails to enable robots, respectively, to return to previously discovered resource locations and to share those locations with other robots. This approach achieved consistent retrieval times when resources were highly clustered; however, the algorithm exhibits degraded performance when resources follow Power-law or random spatial distributions. A substantial improvement over CPFA was reported in \cite{MatthewDDSA2016}, where the authors incorporated a squared-spiral search trajectory to more systematically survey the environment. Although this method yielded superior results, robots in that framework spent a considerable portion of their time traversing a predetermined path, and the peripheral regions of the environment required a prolonged period to be explored. This limitation was subsequently mitigated in \cite{RyanDDSA} by partitioning the environment into \(N\) regions. Nevertheless, all experiments in that work were conducted in relatively small arenas; consequently, as the foraging area increases, the number of regions also grows, which may lengthen the time needed to collect resources when they are concentrated within only a subset of those regions. 

In~\cite{2016IROSMPFA, 2019LuPhD, Lu2018DynamicDepots}, the authors proposed a multiple–placed foraging algorithm that further improves foraging performance. In this approach, multiple collection zones or multiple dynamic depots are deployed throughout the environment, and robots deliver resources to their nearest depot. The depots then relocate to the centroid of the resource locations discovered by the robots. By distributing multiple depots in this manner, our method alleviates congestion around the single central collection zone used in CPFA and reduces the average travel distance required for robots to deliver resources.

Furthermore in the Grid-Based CPFA (GPFA) \cite{arturo2025}, we proposed a way to separate the map into grid sections, then kept track of the number of times a grid was visited and sent robots to the least visited locations in 3$\times$3 grid, once there the robot would perform a spiral search similar to \cite{RyanDDSA}, but as CPFA the algorithm was wasting about 20\% of its time looking for the last 10\% of resources.


%
\section{The Central Placed Foraging Algorithm for Robot Swarms}
\label{cpfa}

\begin{figure}
    \centering
    \includegraphics[width=0.55\textwidth]{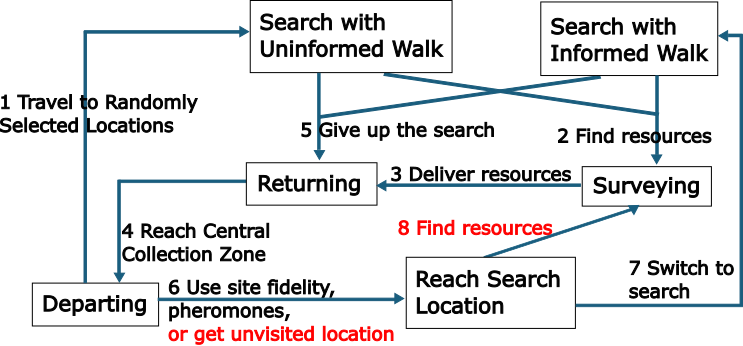}
    \caption{The flow chart of an individual robot’s behavior and states in the CPFA; additions to the algorithm for ARPC are shown in red.}
    \label{fig:CPFA}

\end{figure}

To establish the theoretical control for our study, we utilize the canonical foraging algorithm, CPFA~\cite{BeyondPherom2015}. The robot controller is a finite state machine that governs the transitions between traveling, searching, surveying, and returning to a central
collection zone as described in~\cite{BeyondPherom2015} (see Fig.~\ref{fig:CPFA}); changes to the original algorithm for our approach are shown in red.

\vspace{-6mm}

\paragraph{\textbf{Departing:}}
The robots depart from the central collection zone to a randomly
selected location. Robots have the probability to stop on the way at any time
and transition to Searching.
Robots do not have any prior information about the location of resources
in the environment. Every robot can remember the location of a previously found
resource and can return to this location, a process called site fidelity in ants \cite{Beverly01052009}.
The robot cannot remember all previously visited locations and can only remember the location in the last round of searching. Robots can also communicate
using pheromones \cite{Jackson2007, Pheromone2003}, which are simulated as artificial waypoints \cite{Campo2010} to recruit robots to known locations with resources. If a robot returns (Returning) to the center and then departs, it can depart using site fidelity or pheromone waypoints in the center.
\paragraph{\textbf{Searching:}} Robots search for resources using random walk \cite{Crist1991}. Successful searches lead to Surveying. If a robot finds resources, it can only pick up one resource and deliver it to the center. Otherwise, it keeps searching, and it has the probability of giving up the search and returning to the center, which leads to Returning.
\paragraph{\textbf{Surveying:}} Robots assess the local resource density within a search radius, recording the number of resources. The density information will be used by the central server in the center to determine the creation of pheromone waypoints \cite{BeyondPherom2015}.
\paragraph{\textbf{Returning:}} Robots deliver resources back to the center. The density of resources ${\lambda}_{lp}$ is taken into account to generate a probability of laying a new pheromone waypoint. The robots then restart the cycle from Departing.
The exploration strategy, "Uninformed Search," is triggered when a robot
has no prior information about resource locations (pheromone or site fidelity)
and is unable to exploit them. In this state, the robot selects a random location close
to the boundaries of the arena. While traveling toward this target location, the
robot has a probability (76\% every 5 seconds) of transitioning into a Correlated
Random Walk (CRW). This CRW is a stochastic movement pattern where the
robot’s heading is updated by a random angle and the step size is fixed.


%
\section{Methodology}
\label{methods}

The proposed method, ARPC, maintains a persistent, incrementally updated set of spatial clusters defined over the locations visited by the robots, where they place repulsive pheromones to avoid visiting that location during informed searches. The procedure comprises four stages and is governed by four tunable parameters: the distance tolerance between recorded locations, $\alpha$; the recording frequency, $\rho$; the maximum number of recorded locations per robot, $\lambda$; and the growth tolerance of cluster size, $\tau$.

One of the primary limitations of the original CPFA algorithm is its tendency to leave large regions of the environment unexplored for extended periods of time. In contrast, the improved GPFA algorithm assigns equal priority to all locations within the environment, relying solely on visitation frequency within a fixed 3$\times$3 grid and thus ignoring other potentially informative spatial or contextual features. 

We propose an enhancement of the CPFA exploration phase that enables (i) rapid elimination of regions that do not warrant further exploration, (ii) dynamic identification of high-value locations for subsequent visits, and (iii) early initiation of informed search behaviors. The objective is to ensure that the environment is explored in a more consistent, comprehensive, and efficient manner. 

\subsection{Clustering}

The proposed algorithm modifies the CPFA controller in three principal ways. While in the \textsc{surveying} state, each robot maintains a record of repulsive-pheromone waypoints, up to a maximum capacity of $\lambda$ entries. Once this capacity is reached, newly generated repulsive pheromones overwrite the oldest entries in the list. Simultaneously, the robot’s probability of transitioning to the \textsc{returning} state is increased.

A robot communicates its recorded repulsive pheromones to the nest only when it returns with a resource or when it terminates its search without success. At this point, the robot transitions to the \textsc{departing} state and selects one of the following four actions (cf. Fig.~\ref{fig:CPFA}). 1) Return to the most recent successful collection site (site fidelity); 2) Follow a pheromone trail left by another robot; 3) Choose a random location outside all repulsive-pheromone clusters. The robot moves to this target while logging visited positions and scanning for resources. If it detects a resource en route, it immediately returns to the Central Collection Zone. If none are found upon arrival, it performs a spiral search centered at that location with radius $\tau$; 4) Perform an undirected random search.

Options (1) and (2) are chosen using probabilities from the Poisson cumulative distribution function. Option (3) is the complementary event, selected when the probabilities for (1) and (2) are not high enough.

Selection operates on clusters of locations rather than individual robot locations, reducing computational cost.

The system’s operational states and transition rules match those of the CPFA controller, with one key difference. In ARPC, when robots return to the nest and no pheromone trail or site-fidelity marker is available, they may initiate an informed search guided by repulsive pheromones instead of defaulting to random search. The nest generates up to 1000 random candidate locations and selects the first one outside all repulsive-pheromone clusters. The robot travels to this target and performs a spiral search of radius \(\tau\). If this search finds no resource, the robot switches to random search until it finds a resource or its memory is full.

The nest-level algorithm was modified to include and process repulsive-pheromone data reported by all robots in the simulation. Upon receiving these data, the nest runs the clustering Algorithm~\ref{algo:cluster} in four stages: (i) newly reported locations are incorporated into existing clusters, increasing their visit counts and reinforcing the inference that these regions have already been explored; (ii) DBSCAN \cite{DBSCAN1996} is applied to group locations within an $\epsilon$-neighborhood, where $\epsilon$ is defined as $\alpha$ plus a small offset to prevent immediate clustering of locations from the same robot; (iii) clusters whose extents overlap within the previously defined $\epsilon + radius$ are merged only if Equation~\ref{eq:1} holds, where $A_{i}$ is the area of the $i$-th cluster and $\tau$ is a threshold set to 1\% of the area of the current arena. This entire procedure is executed by the nest at fixed 5-second intervals.

\begin{equation}
\sum_{i} A_{i} \;-\; \sum_{i < j} A_{i \cap j}
\;<\; \tau \, \alpha \, \sum_{i} v_{i}
\label{eq:1}
\end{equation}

\vspace{-4mm}
\begin{algorithm}
\caption{Update Repulsive Pheromone Clusters}
\scriptsize
\begin{algorithmic}[1]
\Require $RepulsivePheromones$ (new), $ExistingRepulsivePheromones$ (memoized singletons),  $VisitedClusters$ (memoized clusters)
$\epsilon \gets \alpha$ + c, $growthSlack \gets \epsilon$
\Statex

\Comment{\textbf{Step 1: Greedy absorption of singletons into existing clusters}}
\For{$p \in RepulsivePheromones \cup ExistingRepulsivePheromones$}
    \If{$\exists c \in VisitedClusters$ such that $Distance(p, c.center) \leq c.radius$}
        \State $\text{Merge}(c, p)$
        \State Remove $p$ from its respective list
    \EndIf
\EndFor
\Statex

\Comment{\textbf{Step 2: DBSCAN on remaining new locations}}
\State $NewClusters \gets \text{DBSCAN}(RepulsivePheromones, \epsilon, minPts=2)$
\State $UnclusteredNew \gets RepulsivePheromones \setminus \{ \text{points in } NewClusters \}$
\Statex

\Comment{\textbf{Step 3a: Merge new unclustered pheromones existing clusters}}
\For{$p \in UnclusteredNew$}
    \If{$\exists ep \in ExistingRepulsivePheromones$ within $\epsilon$ distance of $p$}
        \State $c_{new} \gets \text{CreateCluster}(p, ep)$
        \State Add $c_{new}$ to $VisitedClusters$
        \State Remove $ep$ from $ExistingRepulsivePheromones$
    \Else
        \State Add $p$ to $ExistingRepulsivePheromones$
    \EndIf
\EndFor
\Statex

\Comment{\textbf{Step 3b: Merge new clusters with existing clusters}}
\For{$nc \in NewClusters$}
    \If{$\exists c \in VisitedClusters$ such that $nc$ overlaps with $c$}
        \State $\text{Merge}(c, nc)$
    \Else
        \For{$ep \in ExistingRepulsivePheromones$}
            \If{$ep$ falls within $nc.radius + growthSlack$}
                \State Add $ep$ to $nc$
                \State Remove $ep$ from $ExistingRepulsivePheromones$
            \EndIf
        \EndFor
        \State Recompute $nc.center$ and $nc.radius$
        \State Add $nc$ to $VisitedClusters$
    \EndIf
\EndFor
\Statex

\end{algorithmic}
\label{algo:cluster}
\end{algorithm}

\begin{figure}[t!]
    \centering
    \begin{subfigure}[b]{0.4\columnwidth}
        \centering
        \includegraphics[width=\linewidth]{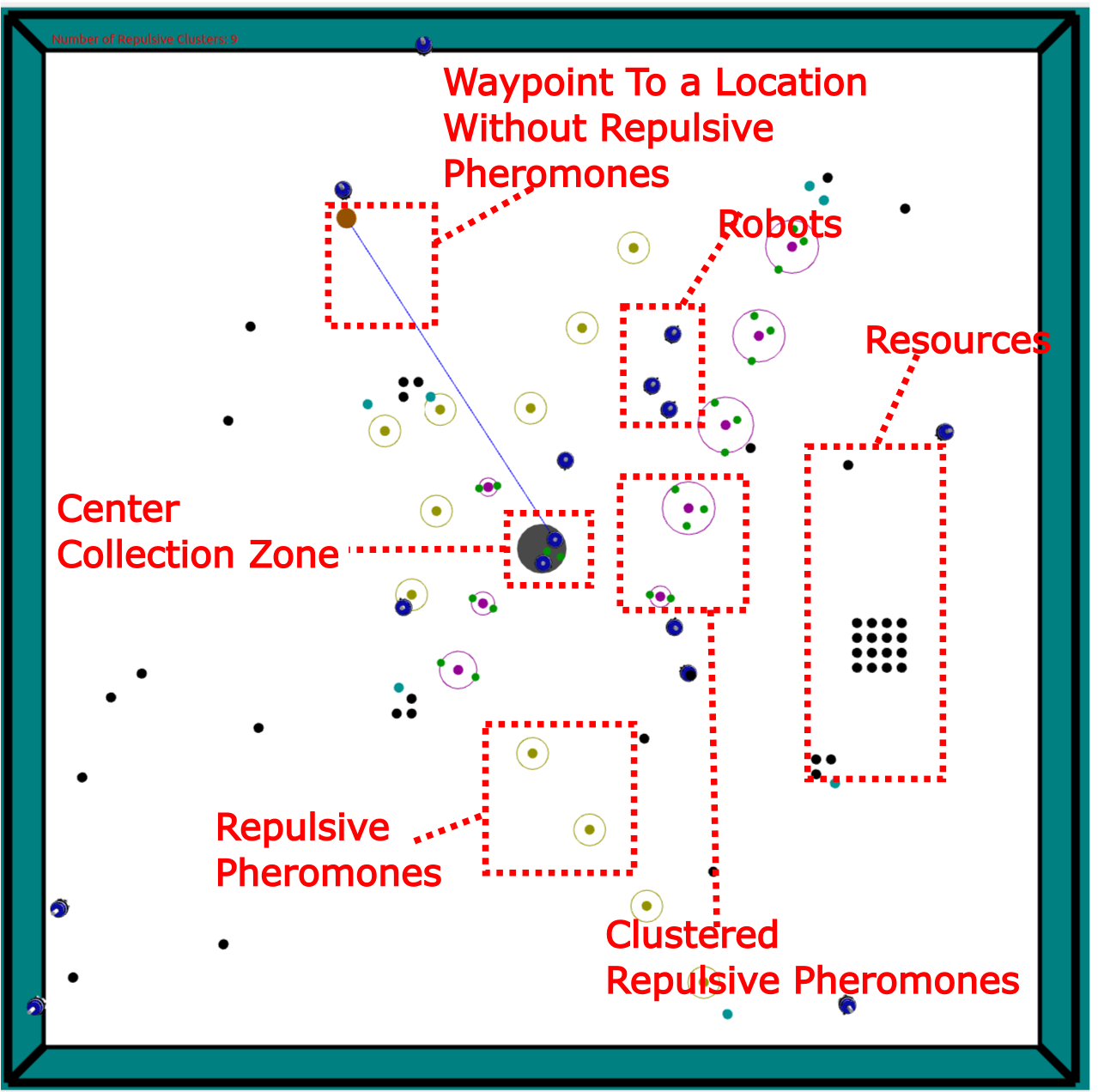}
        \caption{}
    \end{subfigure}

    \par\medskip

    \begin{subfigure}[b]{0.24\columnwidth}
        \centering
        \includegraphics[width=\linewidth]{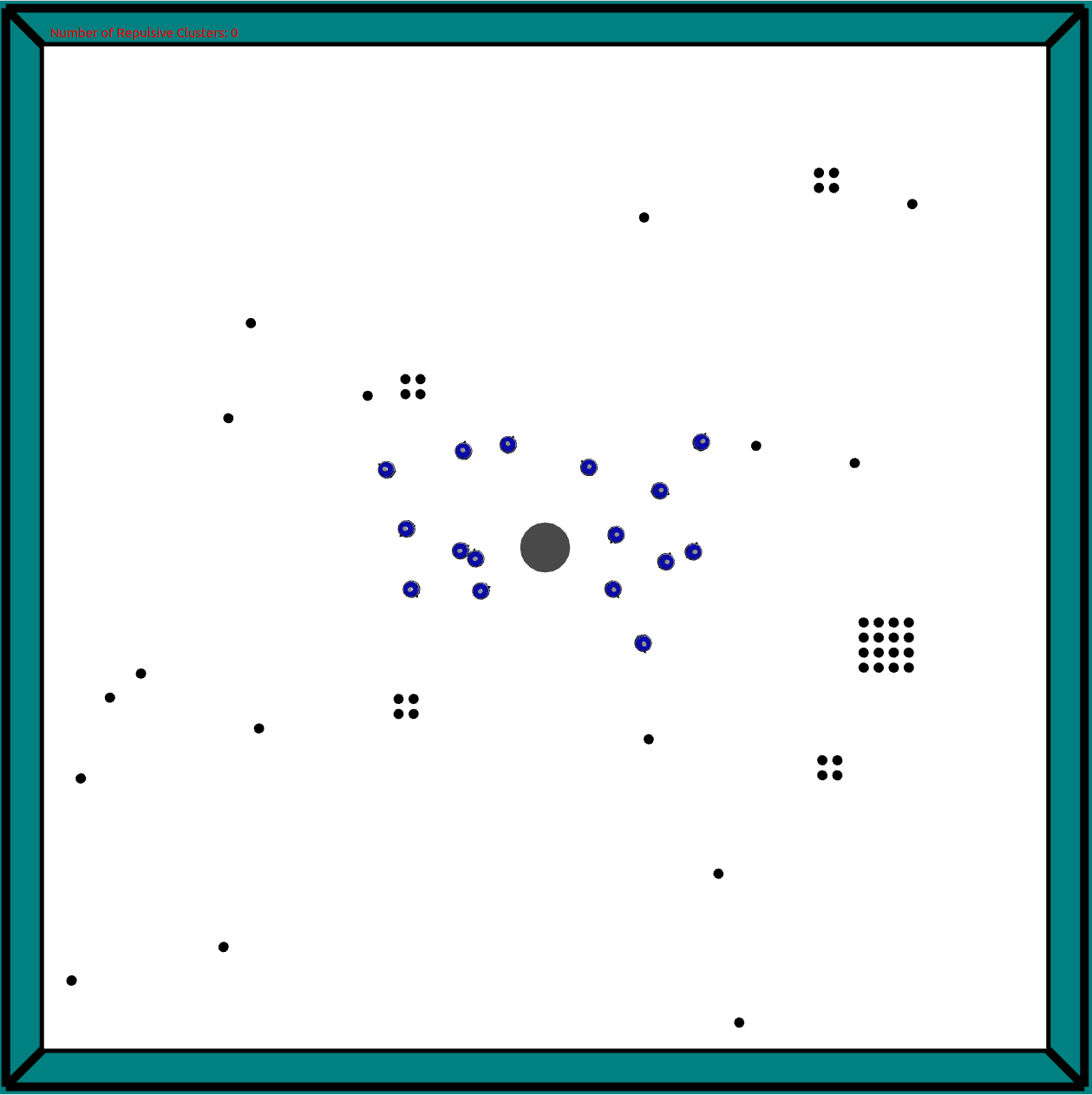}
        \caption{0.25 min}
    \end{subfigure}\hfill
    \begin{subfigure}[b]{0.24\columnwidth}
        \centering
        \includegraphics[width=\linewidth]{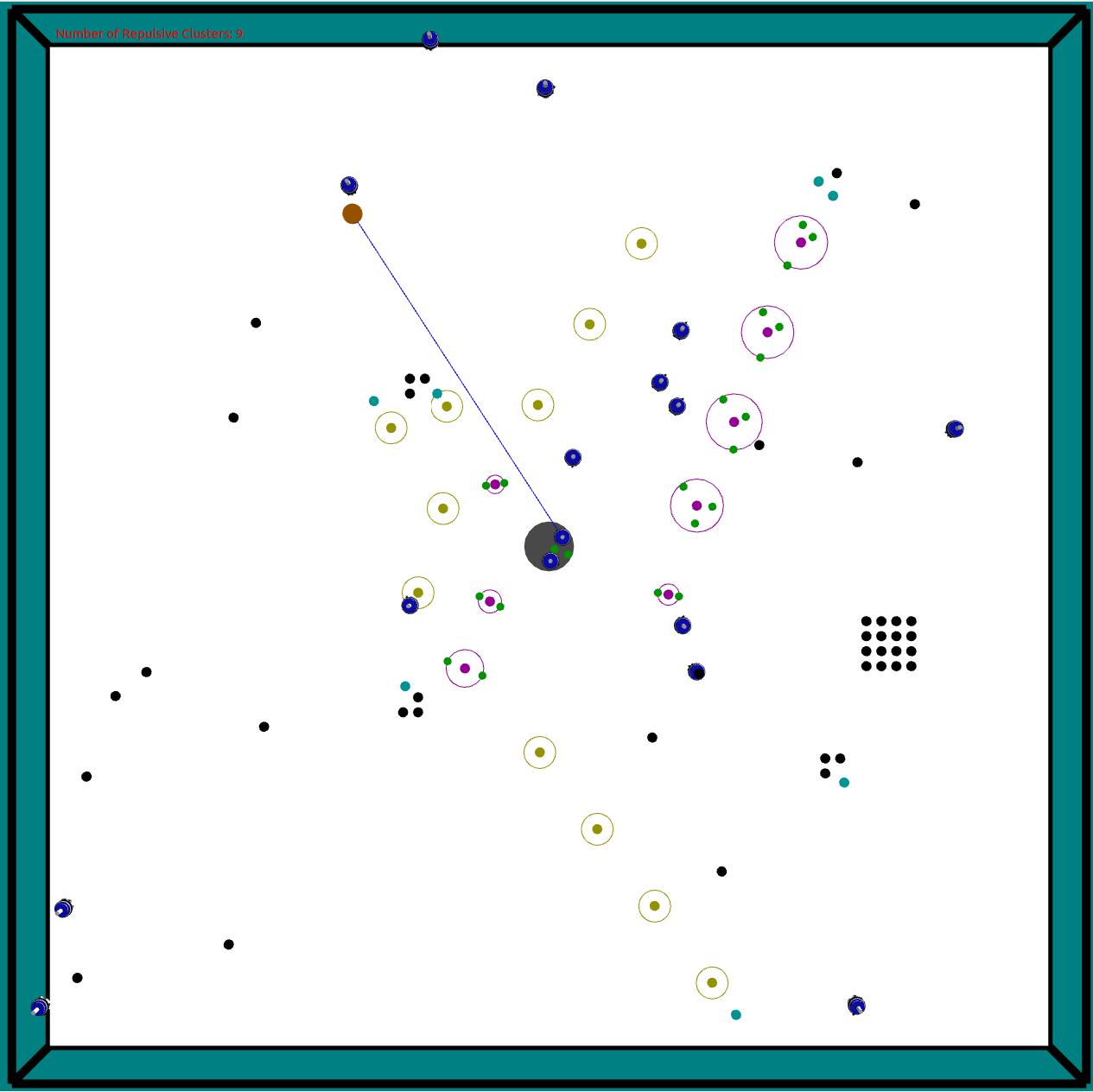}
        \caption{1.25 min}
    \end{subfigure}\hfill
    \begin{subfigure}[b]{0.24\columnwidth}
        \centering
        \includegraphics[width=\linewidth]{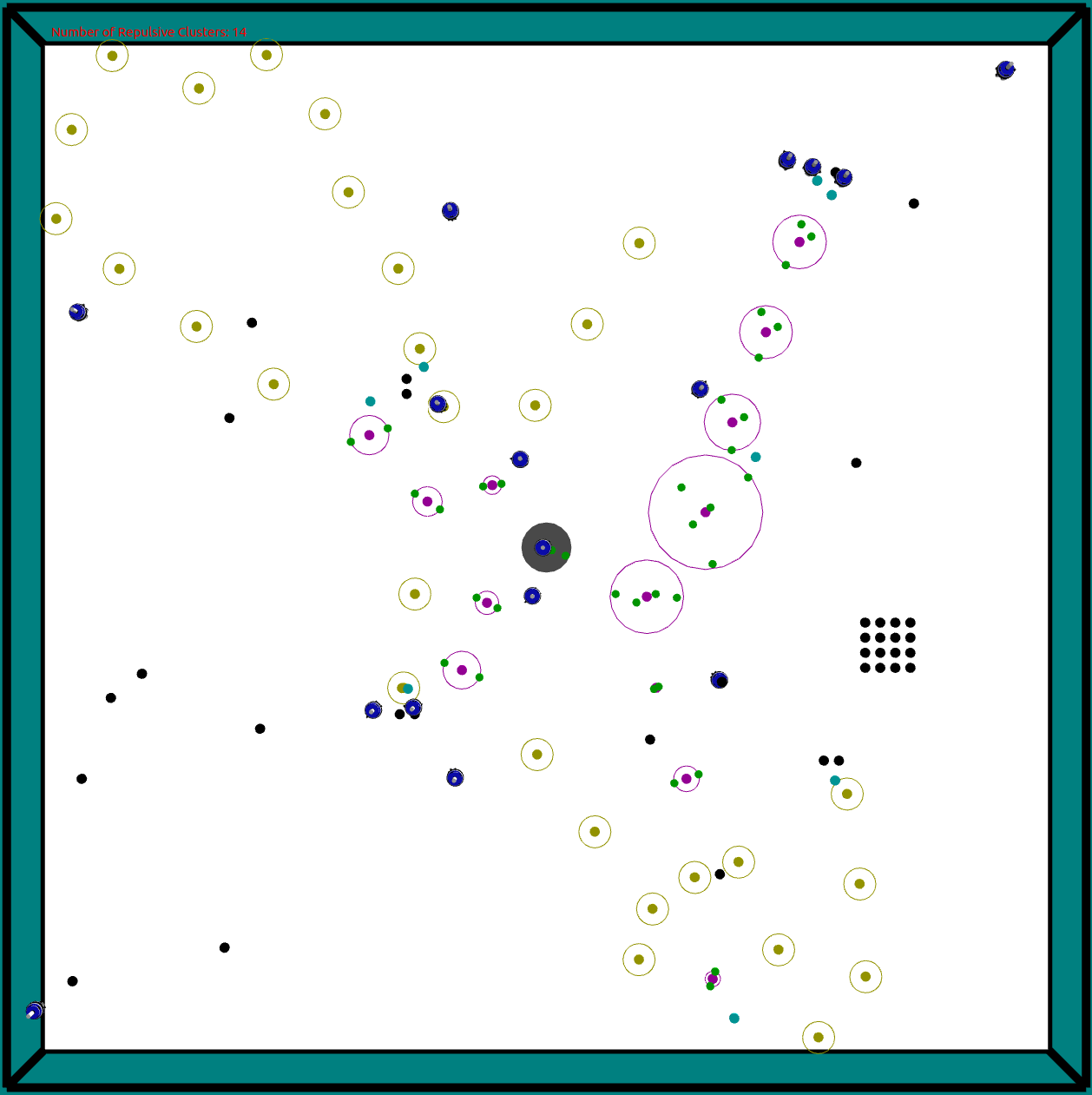}
        \caption{2 min}
    \end{subfigure}\hfill
    \begin{subfigure}[b]{0.24\columnwidth}
        \centering
        \includegraphics[width=\linewidth]{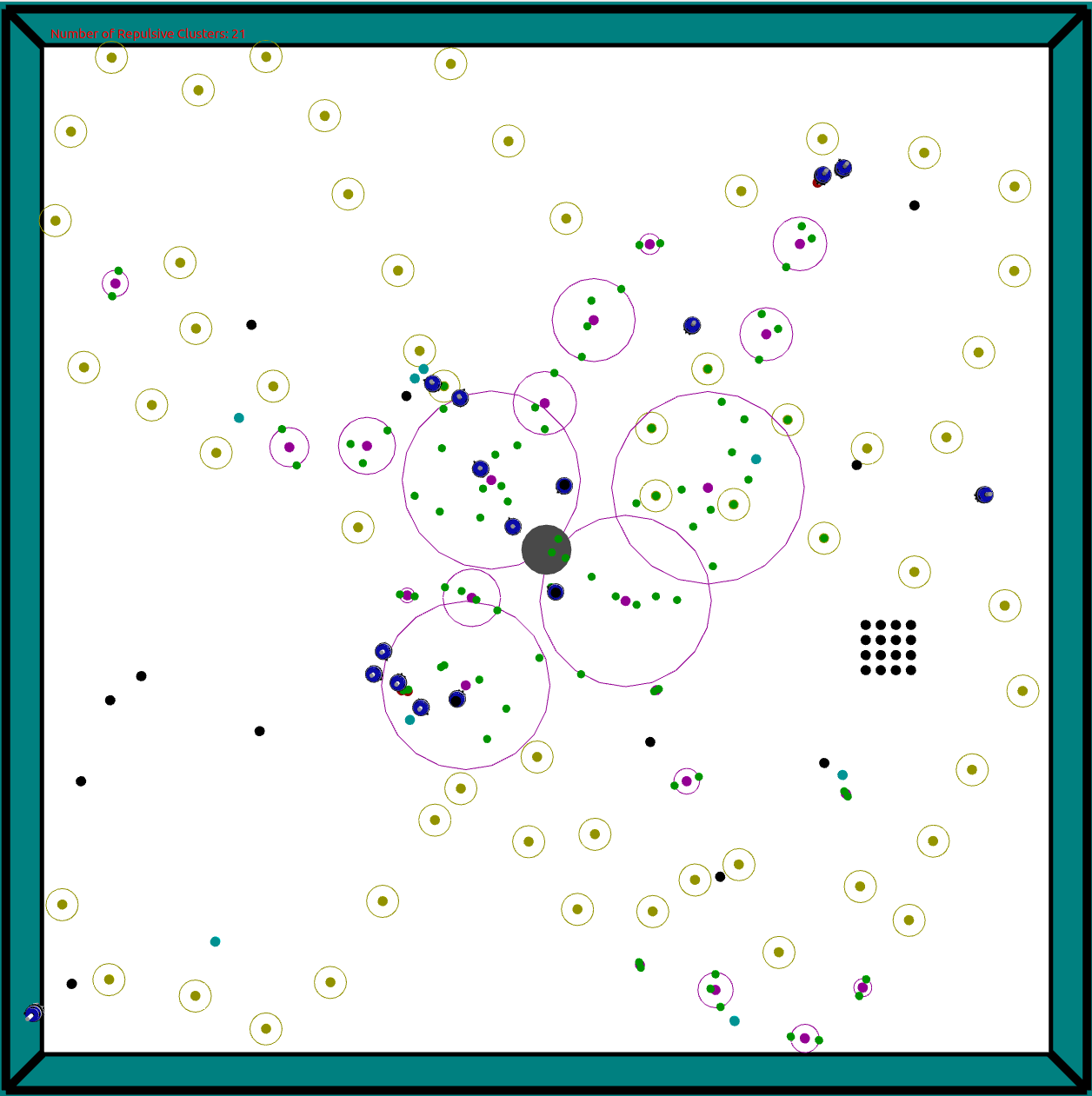}
        \caption{2.75 min}
    \end{subfigure}

    \vspace{0.5ex}

    \begin{subfigure}[b]{0.24\columnwidth}
        \centering
        \includegraphics[width=\linewidth]{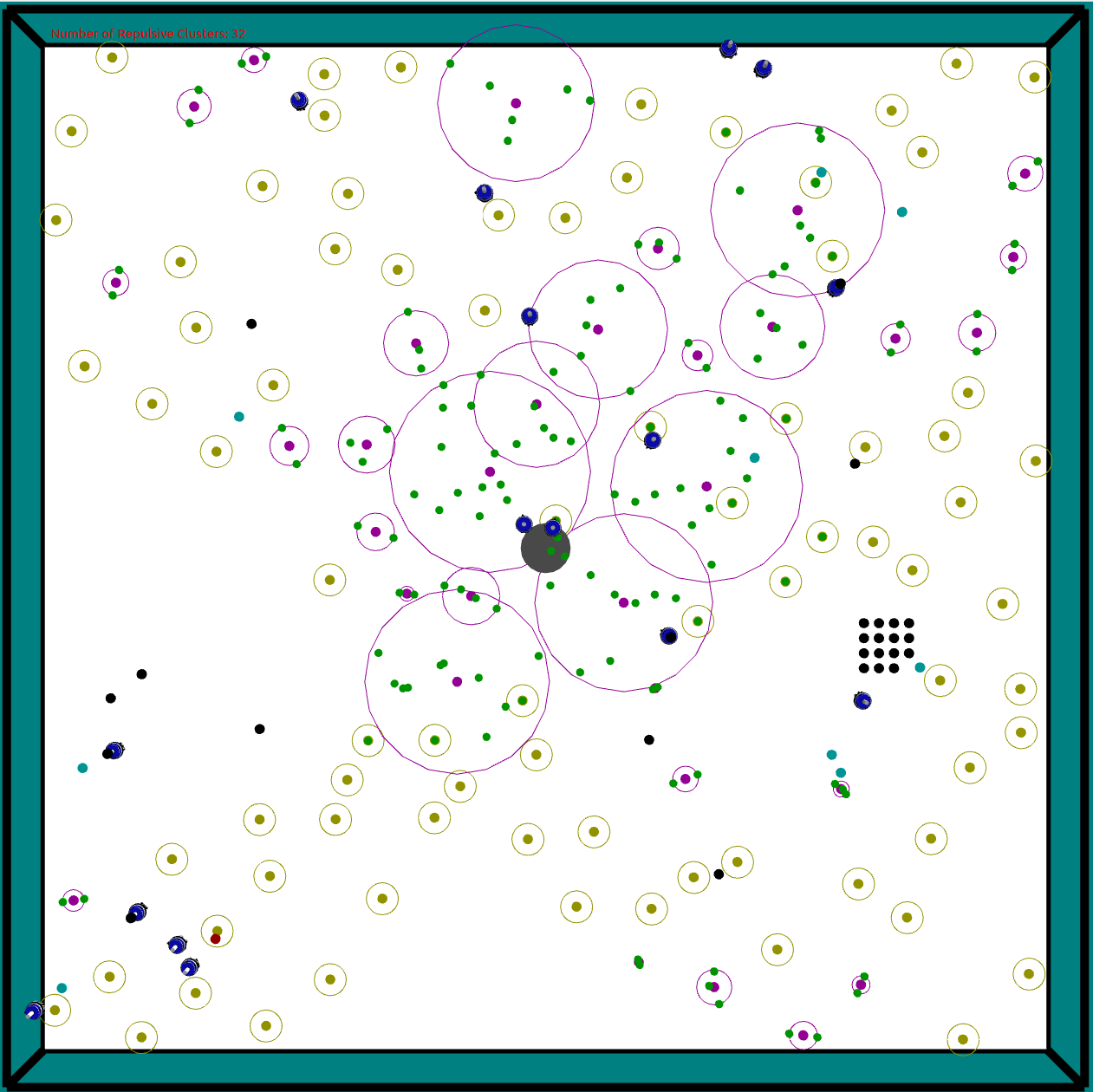}
        \caption{5.75 min}
    \end{subfigure}\hfill
    \begin{subfigure}[b]{0.24\columnwidth}
        \centering
        \includegraphics[width=\linewidth]{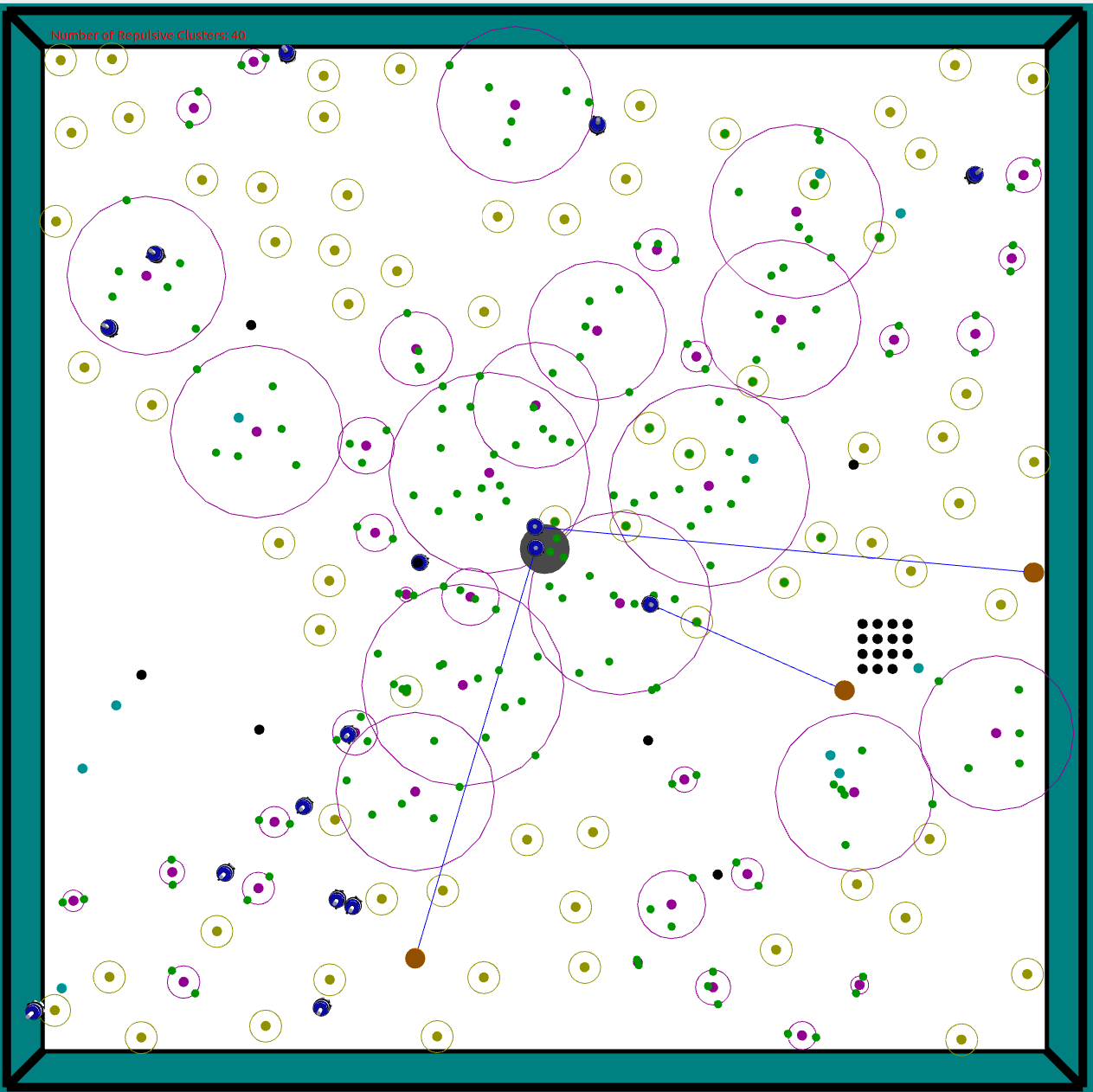}
        \caption{6.25 min}
    \end{subfigure}\hfill
    \begin{subfigure}[b]{0.24\columnwidth}
        \centering
        \includegraphics[width=\linewidth]{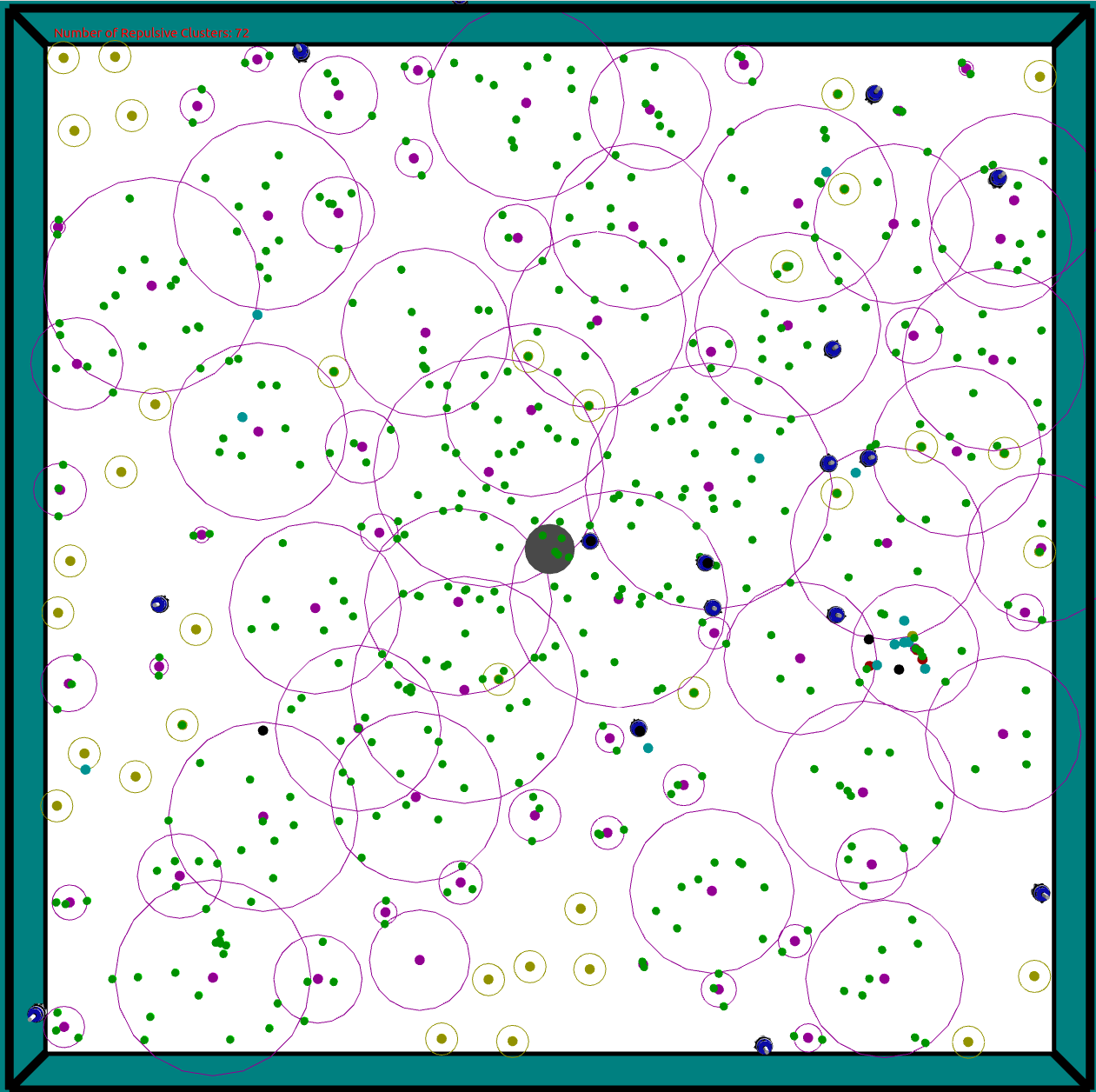}
        \caption{10 min}
    \end{subfigure}\hfill
    \begin{subfigure}[b]{0.24\columnwidth}
        \centering
        \includegraphics[width=\linewidth]{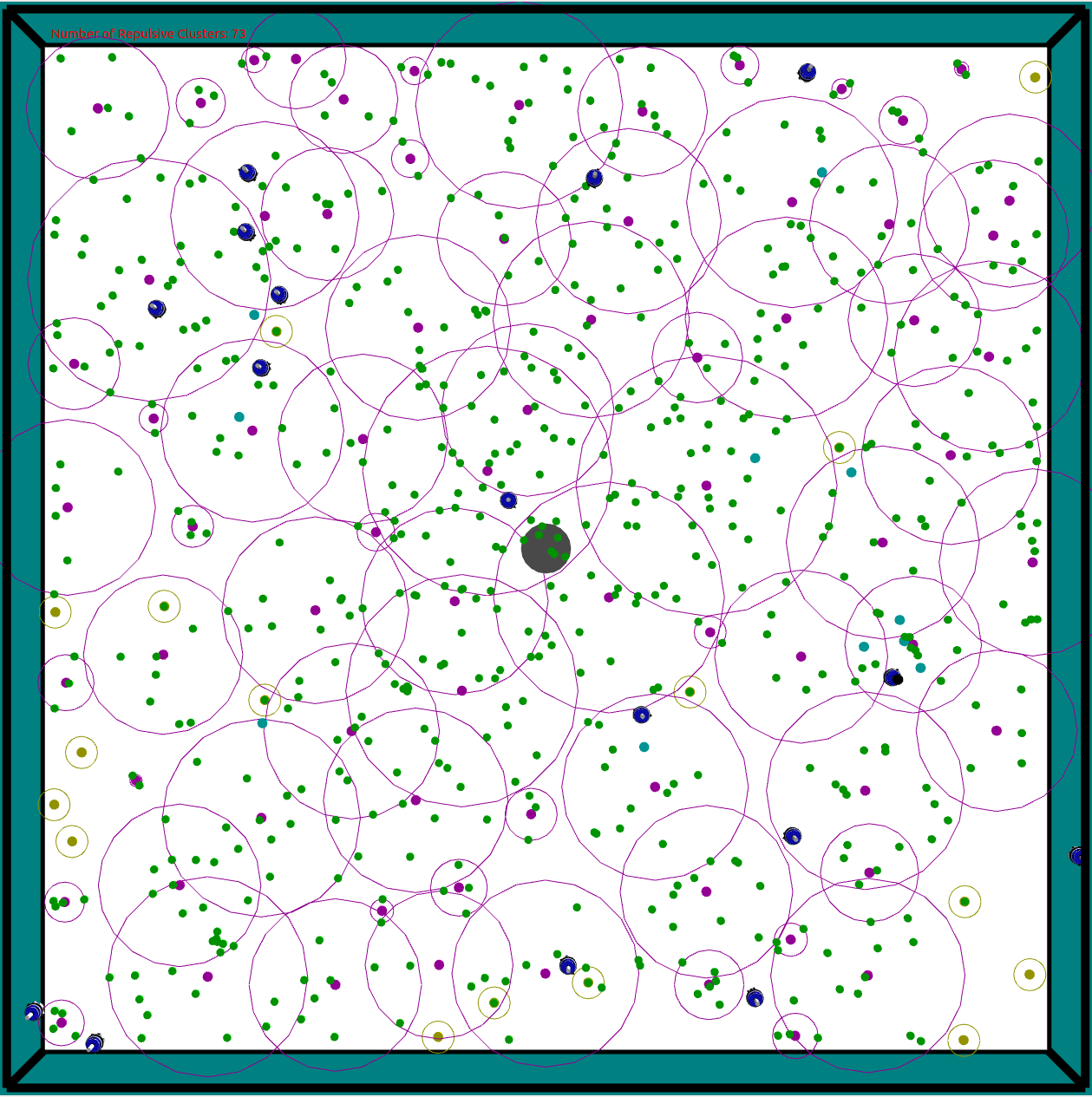}
        \caption{16.25 min}
    \end{subfigure}
    \caption{Example ARGoS simulations conducted in a 10$\times$10 m arena. Panel (a) shows the baseline setup, in which robots explore the environment and gather resources. The robots lay down repulsive pheromones (yellow dots), which later coalesce into clusters (magenta); green dots indicate the individual locations that contribute to each cluster. Panels (b - i) present an entire simulation run.}
    \label{fig:simulation1}
\end{figure}

Fig. \ref{fig:simulation1} illustrates the visualization of repulsive pheromones (yellow) and clustered repulsive pheromones (magenta) within the simulation environment, green dots are used to represent the locations that were used to form the cluster, but do not contribute to the algorithm. Repulsive pheromones are deposited along the trajectory of each robot, and, as the simulation advances, the environment is progressively represented through these pheromone clusters. Based on this emerging pheromone map, the nest subsequently allocates robots to previously unvisited regions, as depicted in Fig. \ref{fig:simulation1}(b), note repulsive pheromones not in a cluster will not be considered in the informed search selection to avoid long computation times. An additional video playlist is available on our YouTube \footnote{\url{https://tinyurl.com/clusteringForaging}}.


\vspace{-6mm}

\section{Experiments}
\label{experiments}

We employed the ARGoS multi-robot simulator to assess the effectiveness of ARPC \cite{argos}. The experimental scenarios were originally proposed by Arturo et al. \cite{arturo2025} and are designed to evaluate the foraging performance of our method across a variety of unknown environments. Foraging performance is quantified as the time required to collect all available resources. The experimental design emphasizes repeatability and reliability.
We performed three distinct experiments to evaluate foraging performance across different spatial distributions, runtime milestones, and arena sizes. The resource distributions (Clustered, Power-law, and Random) emulate canonical setups \cite{BeyondPherom2015}. All experiments utilized 16 Foot-Bot robots \cite{foot-bot} in arenas that differed either in size or in the number of resources, allowing us to evaluate the algorithm under progressively lower robot-to-resource densities. Each experimental setup was run 50 times to reduce statistical bias.The ARPC algorithm was parameterized using the baseline parameters $\alpha = 0.75$ m, $\rho = 2$ s, $\lambda = 50$, and $\tau = \textit{AreaArena} \times 0.005$. A detailed summary of the experimental setups and associated independent variables for all three experiments is provided in Table \ref{tab:experiments}.

\vspace{-4mm}
\begin{table}[h!]
    \caption{Consolidated Experimental Configurations}
    \centering
    \scriptsize
    \begin{tabular}{|c|c|c|c|}
    \hline
         Experiments &  I & II & III\\ \hline
         Arena Size (meter) & 14 × 14 & 14 × 14 & \begin{tabular}[c]{@{}c@{}} 8 × 8, 10 × 10, 12 × 12 \\ 14 × 14, 16 × 16 \end{tabular} \\ \hline
         \# of Resources & 16, 32, 48, 64, 80 & 48 & 48 \\ \hline
         \# of Robots & 16 & 16 & 16 \\ \hline
         \# of Runs & 50 & 50 & 50 \\ \hline
         Resource Distributions & \begin{tabular}[c]{@{}c@{}} Clustered \\ Powerlaw \\ Random \end{tabular} & \begin{tabular}[c]{@{}c@{}} Clustered \\ Powerlaw \\ Random \end{tabular} & Random \\ \hline
    \end{tabular}
    \label{tab:experiments}
\end{table}
\vspace{-6mm}

\section{Results}
\label{discussion}

The experimental outcomes are analyzed along three principal dimensions: (i) evaluation of the algorithm’s scalability as the available computational resources are increased; (ii) quantitative characterization of the efficiency gains achieved during the final collection phase; and (iii) assessment of the algorithm’s robustness across different arena sizes. The statistical significance of performance differences is assessed using the `ttest\_ind` function from the `scipy.stats` library, which implements the independent-samples t-test. Statistical significance is indicated by asterisks in Figures \ref{fig:experiment1}, \ref{fig:experiment2}, and \ref{fig:experiment3}. The notch in each box plot denotes the 95\% confidence interval of the median; non-overlapping notches between two boxes indicate a statistically significant difference between their medians. Significance levels are encoded as follows: ‘***’ denotes a highly significant difference (p < 0.001), ‘**’ denotes a substantial difference (p < 0.01), ‘*’ denotes a moderate difference (p < 0.05), and ‘-’ indicates that the difference is not statistically significant.

\subsection{Flexibility with Different Numbers of Resources}

\begin{figure*}[t!]
    \centering
    \begin{subfigure}[t]{0.5\textwidth}
        \centering
        \includegraphics[height=1.2in]{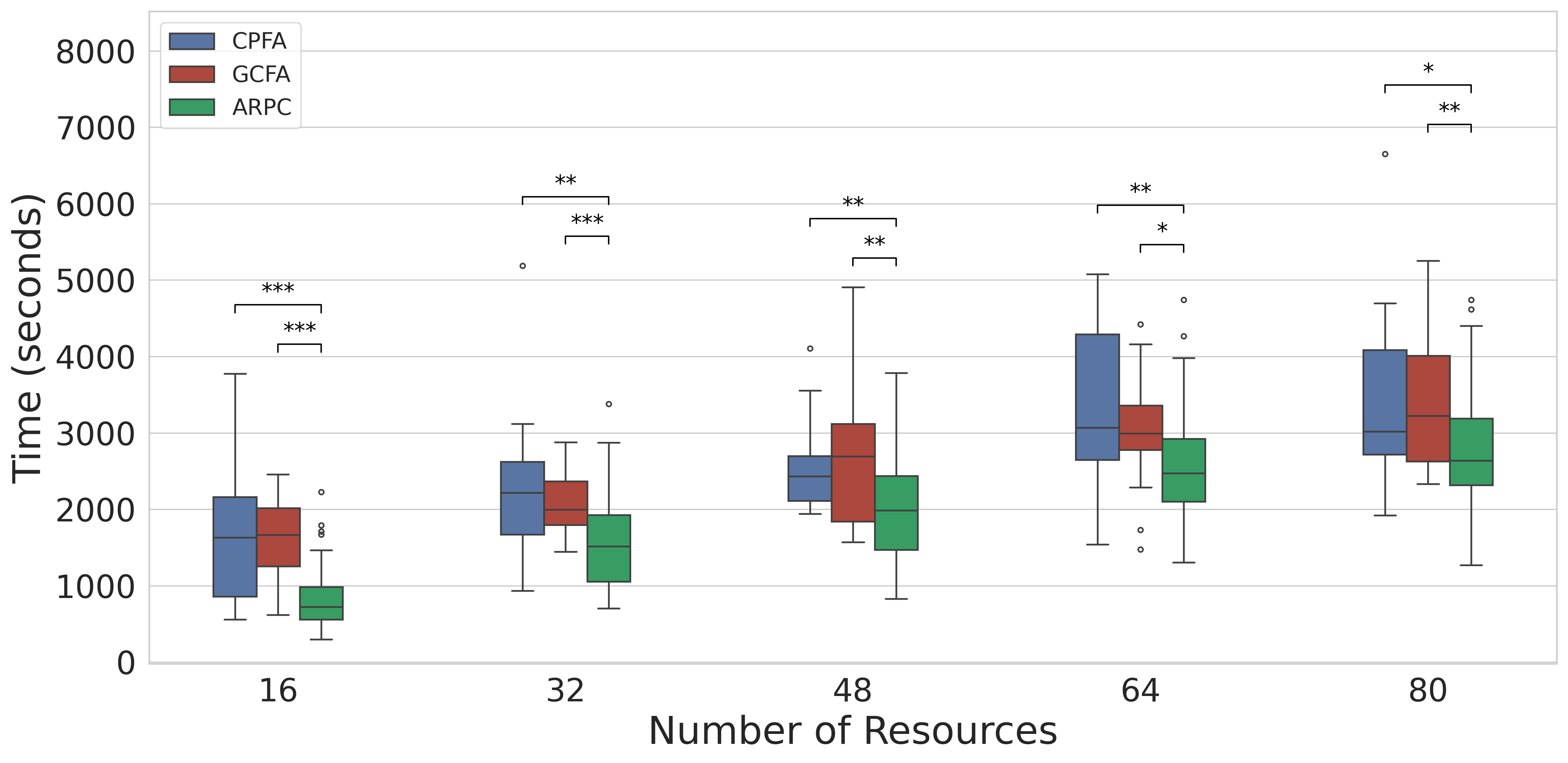}
        \caption{Clustered distribution}
    \end{subfigure}
    ~
    \begin{subfigure}[t]{0.5\textwidth}
        \centering
        \includegraphics[height=1.2in]{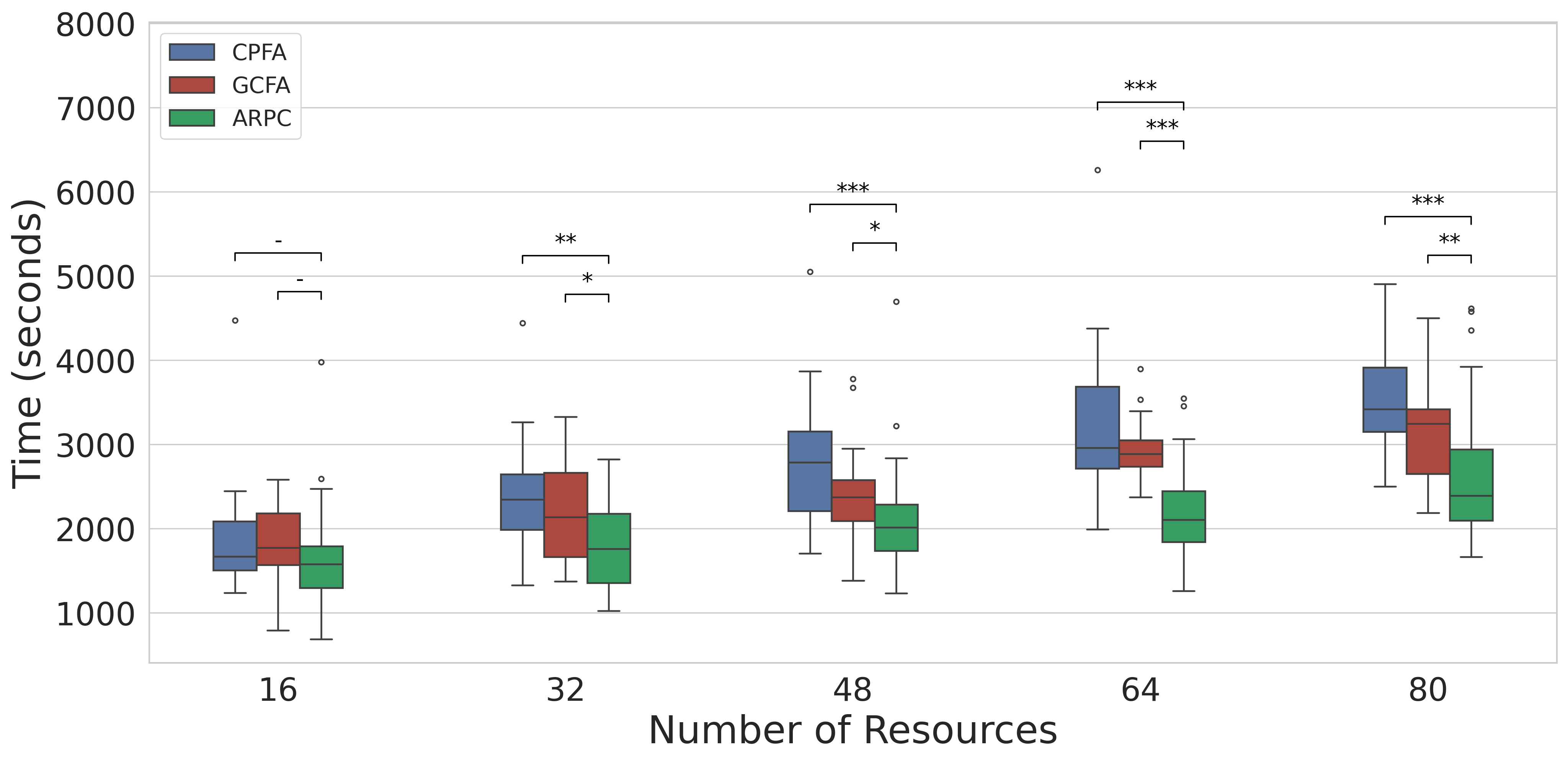}
        \caption{Powerlaw distribution}
    \end{subfigure}%
    ~
    \begin{subfigure}[t]{0.5\textwidth}
        \centering
        \includegraphics[height=1.2in]{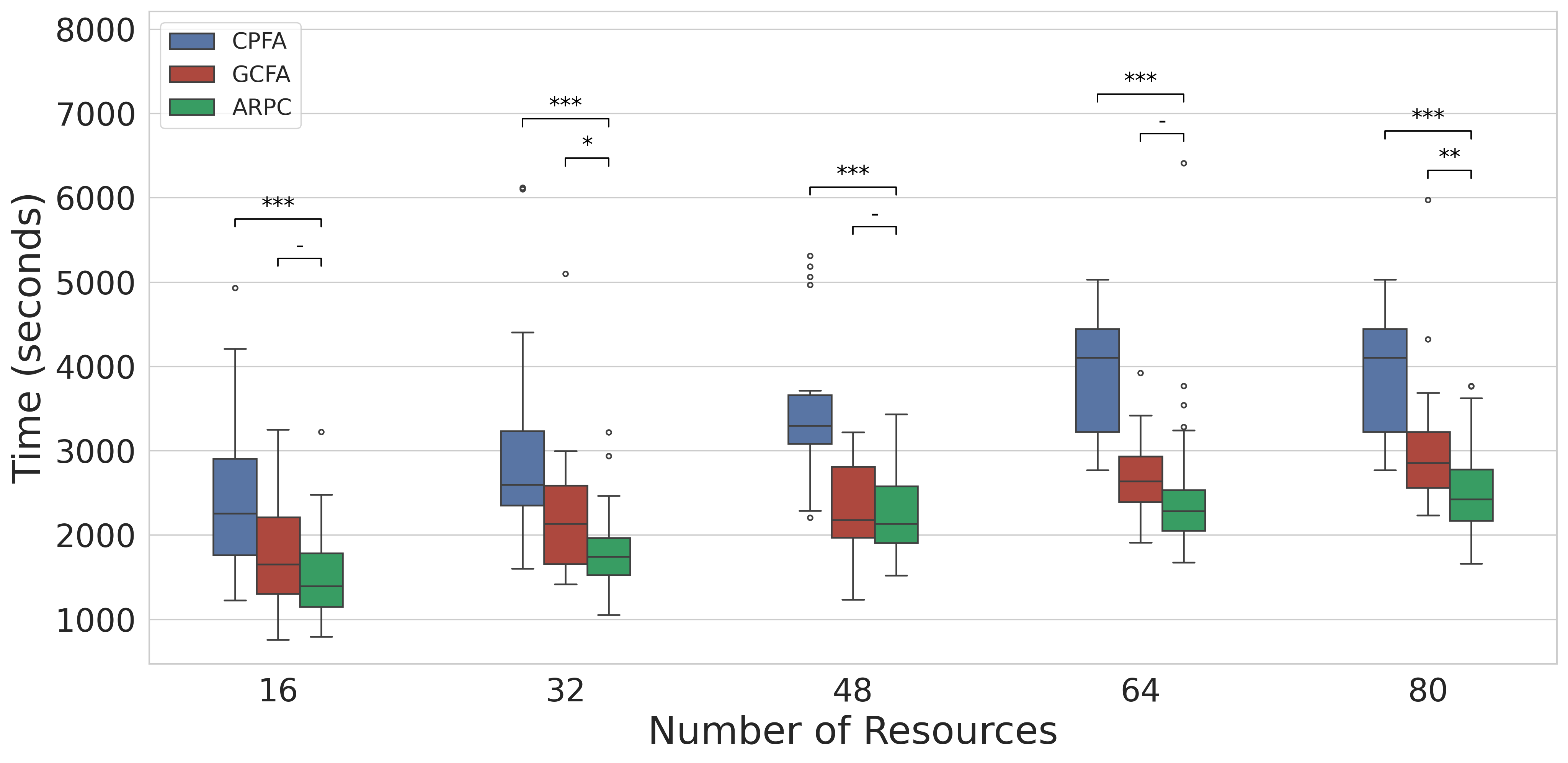}
        \caption{Random distribution}
    \end{subfigure}
    \caption{Experiment I: cumulative time (s) needed to collect 100\% of the resources. All experiments are performed in a 14$\times$14\,m arena containing 16, 32, 48, 64, or 80 resources.}
    \label{fig:experiment1}
\end{figure*}

Figure \ref{fig:experiment1} presents the results of Experiment I. The proposed method (green) consistently outperforms the original CPFA (blue) and GPFA (red) in 8 out of 15 configurations. For clustered resource distributions, our method either matches or exhibits statistically significant improvements over CPFA and GPFA across the range of resource quantities. In exploration-intensive scenarios (Random and Powerlaw distributions), the relative performance of our method varies. Under the Powerlaw distribution, for $N = 64$ and $N = 80$, ARPC yields considerable performance gains. For the Random distribution, GPFA and ARPC exhibit comparable performance for most values of $N$, but at $N = 80$ ARPC demonstrates an improvement, suggesting that the algorithm maintains a relatively stable foraging rate irrespective of the total number of resources present in the environment.

\vspace{-6mm}

\subsection{Analysis of the Final Collection Phase}

\begin{figure*}[htbp!]
    \centering
    \begin{subfigure}[b]{0.5\textwidth}
        \centering
        \includegraphics[width=\textwidth]{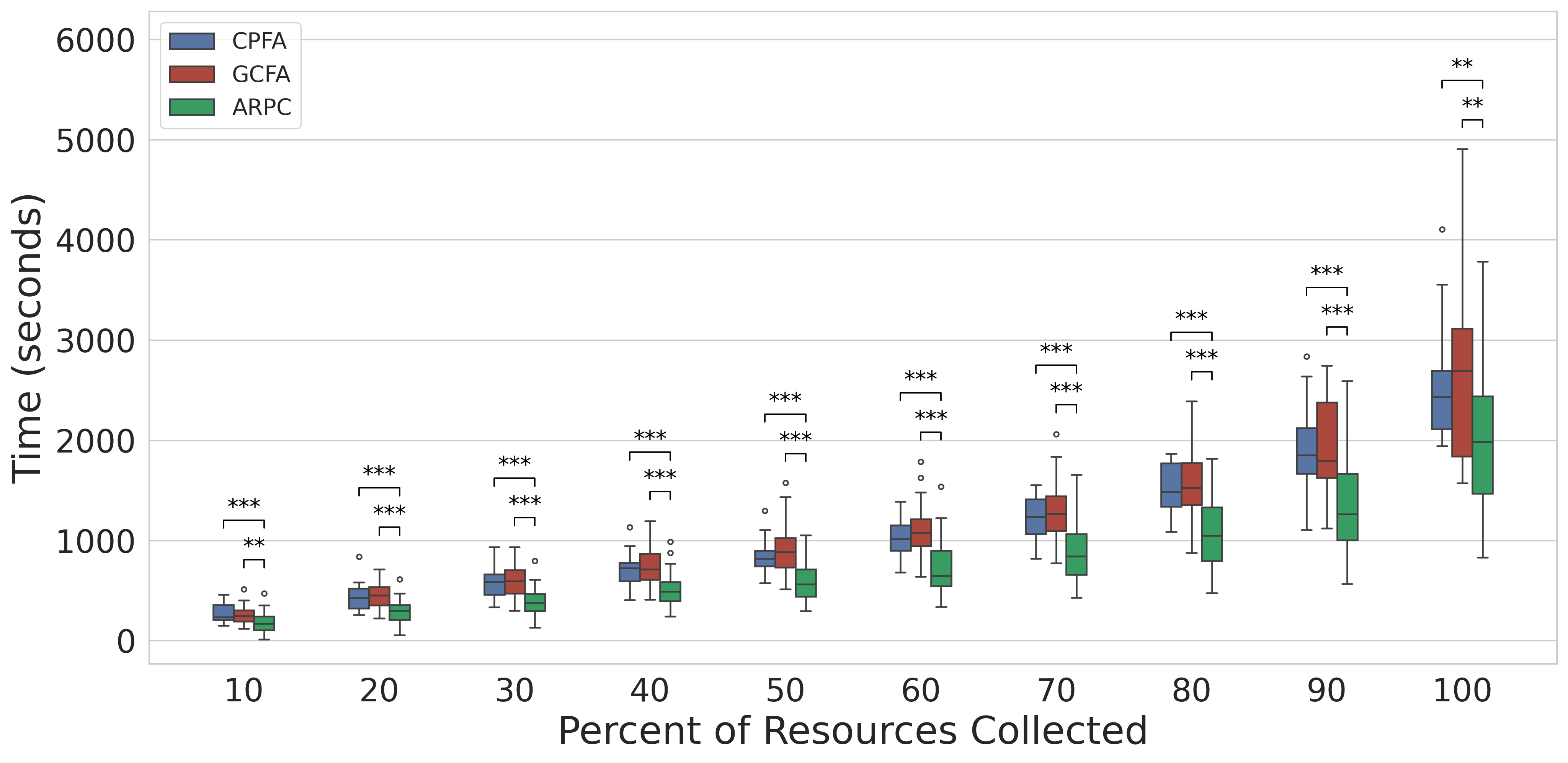}
        \caption{Clustered distribution}
    \end{subfigure}\hfill
    \begin{subfigure}[b]{0.5\textwidth}
        \centering
        \includegraphics[width=\textwidth]{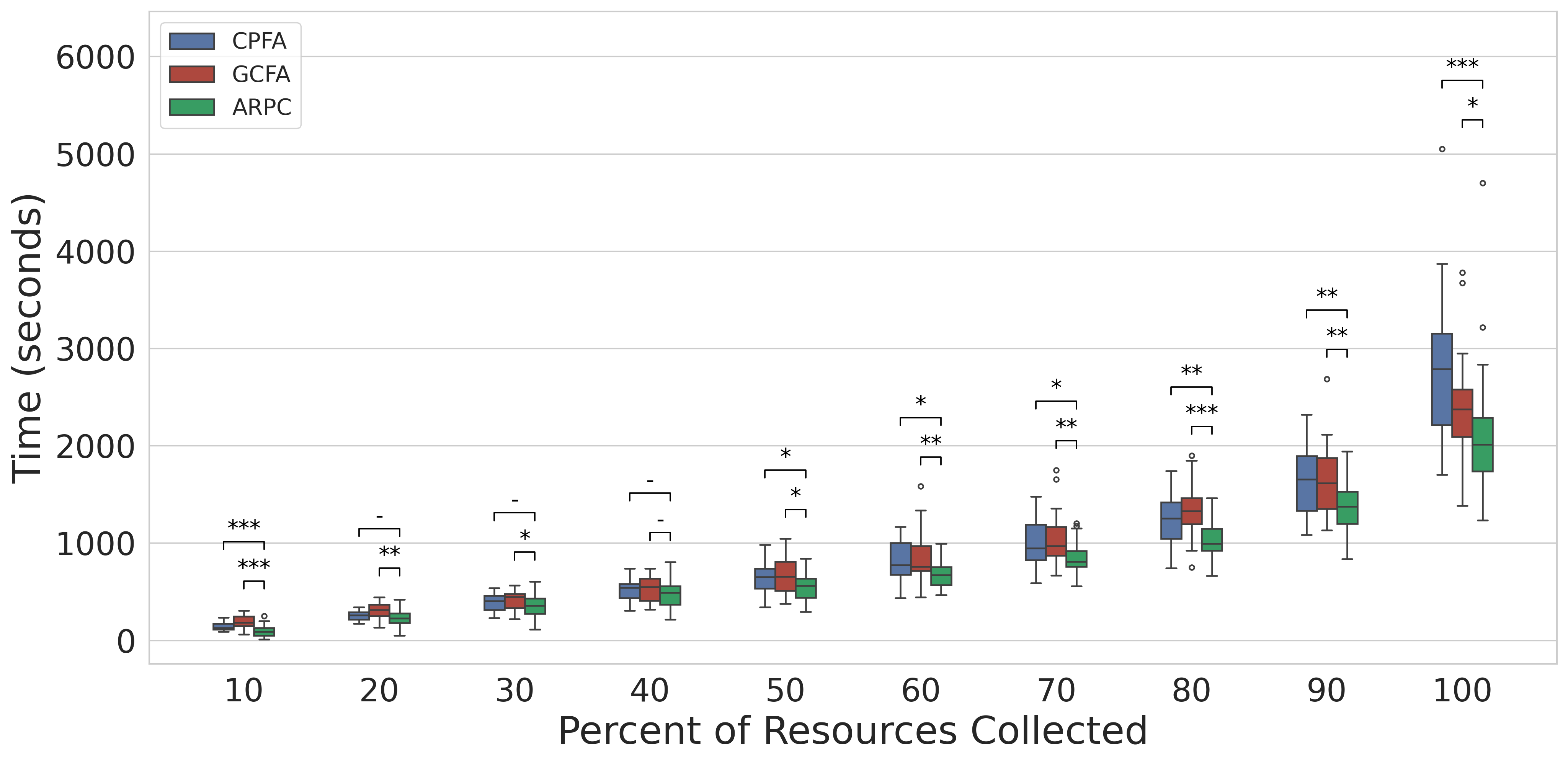}
        \caption{Powerlaw distribution}
    \end{subfigure}\hfill
    \begin{subfigure}[b]{0.5\textwidth}
        \centering
        \includegraphics[width=\textwidth]{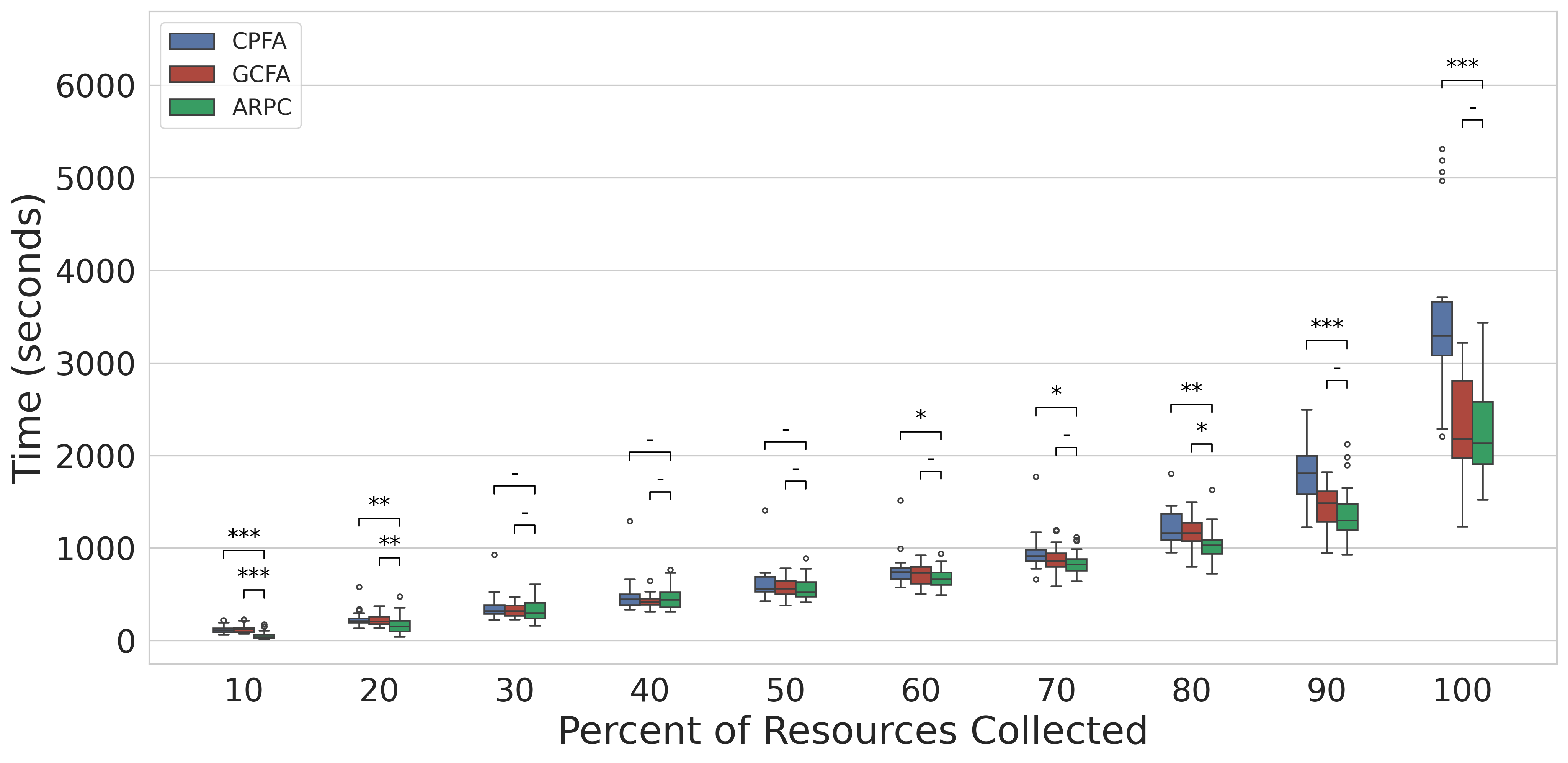}
        \caption{Random distribution}
    \end{subfigure}
    \caption{Experiment II: time (s) to reach each 10\% resource-collection milestones (10\%, 20\%, \dots, 100\%) in a 14$\times$14 arena containing 48 resources, evaluated under three spatial resource distributions: (a) clustered, (b) power-law, and (c) random.}
    \label{fig:experiment2}
\end{figure*}

As previously discussed, CPFA allocates approximately 50\% of its total foraging time to collecting the final 12\% of the available resources. Consequently, in Experiment II we focus on the early and late stages of the foraging process to assess performance differences between the algorithms. Figure~\ref{fig:experiment2} presents the temporal evolution of resource collection under both conditions.

For the Clustered and Powerlaw distributions, beginning at the point where 60\% of the resources have been collected, the proposed algorithm exhibits a substantial performance advantage that increases as the number of remaining resources decreases. Under the Random distribution, our algorithm begins to outperform the baseline once 80\% of the resources have been collected.

To more systematically quantify the efficiency of our method during the final collection phase, we define and analyze two specific intervals: \textbf{Interval A (80\% $\rightarrow$ 90\%):} At this stage, resources are relatively scarce, and locating them becomes increasingly time-consuming. \textbf{Interval B (90\% $\rightarrow$ 100\%):} This final stage corresponds to the regime in which CPFA exhibits its greatest difficulty, expending the majority of its foraging time.

Table~\ref{tab:experiment2} summarizes our results in comparison with GPFA, which was previously reported in \cite{arturo2025} to achieve 49\% and 35\% improvements over CPFA in these two final intervals, respectively.

\begin{table}[!htbp]
    \centering
    \caption{Duration of the Final Collection Phases (Seconds)}
    \scriptsize
    \begin{tabular}{|l|c|c|c|c|c|}
    \hline
       Config  &  \begin{tabular}{@{}c@{}}GPFA\\(80\% $\rightarrow$ 90\%) \end{tabular}&  \begin{tabular}{@{}c@{}}ARPC \\(80\% $\rightarrow$ 90\%)\end{tabular} &  \begin{tabular}{@{}c@{}}GPFA \\(90\% $\rightarrow$ 100\%)\end{tabular} &  \begin{tabular}{@{}c@{}}ARPC \\(90\% $\rightarrow$ 100\%)\end{tabular} & \begin{tabular}{@{}c@{}}Endgame \\Improvement\end{tabular}\\ \hline
       Cluster & 394.0 & 269.3 & 680.7 & 593.8 & \textbf{+12.78\%} \\ \hline
       Powerlaw & 313.7 & 276.9 & 774.8 & 1021.1 & \textbf{$-$31.79\%} \\ \hline
       Random  & 298.6 & 284.0 & 900.8 & 703.7 & \textbf{+21.89\%} \\ \hline
    \end{tabular}
    \label{tab:experiment2}
\end{table}
\vspace{-4mm}

\paragraph{Cluster Distribution:} In this distribution, most of the environment lacks resources, with only a few clustered regions. Rapidly locating these clusters is therefore crucial. Here, GPFA performed 15\% worse than CPFA, whereas ARPC improved on GPFA by 13\% but was 0.03\% below CPFA. ARPC nonetheless outperformed both algorithms at all stages, likely due to its hybrid strategy: CPFA repeatedly revisits the same locations, while GPFA ignores locations once visited within a 3 $\times$ 3 grid. ARPC instead limits redundant revisits without overprioritizing unvisited regions.

\paragraph{Powerlaw Distribution:} In this distribution, many locations contain multiple resources scattered throughout the arena, making them easy to miss. ARPC performed 32\% worse than GPFA, yet still 15\% better than CPFA. ARPC’s marked decline occurs early (10\%–40\% completion; see Fig.~\ref{fig:experiment2}), when it is primarily depositing repulsive pheromones. Isolated resources near dense clusters can fall inside repulsive zones, lowering their likelihood of collection.

\paragraph{Random Distribution:} Here, all resources appear as isolated items across the environment, increasing the chance they are missed unless robots thoroughly survey the area. In the late stage, ARPC outperforms GPFA and CPFA by 22\% and 60\%, respectively. This suggests ARPC effectively removes unnecessary search regions and rapidly directs robots toward areas with higher expected resource density.

\subsection{Robustness Across Arena Scales}

\begin{figure}
    \centering
    \includegraphics[width=0.6\linewidth]{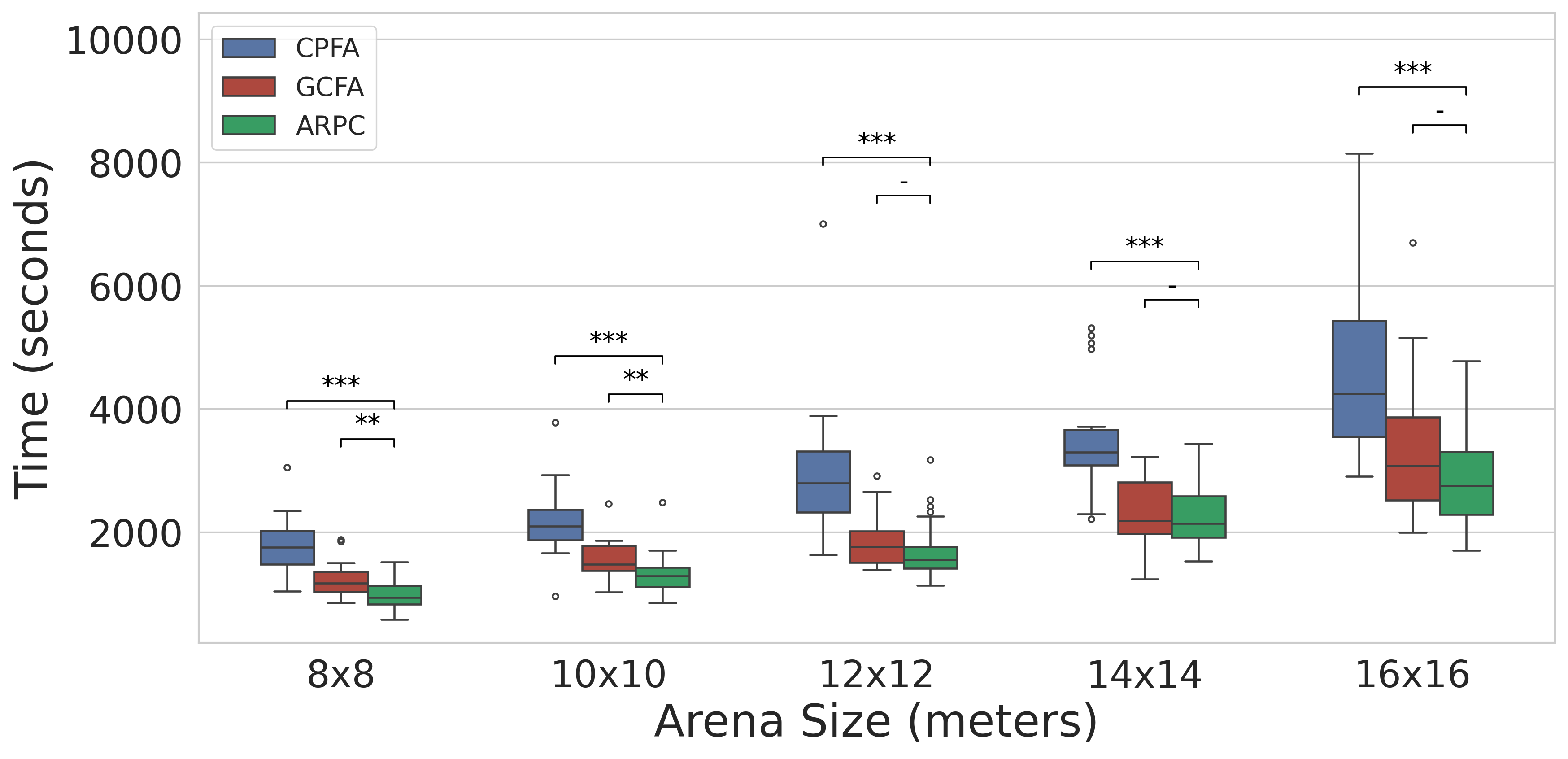}
    \caption{Experiment III: cumulative completion time (s) to collect 100\% of resources across different square arena sizes (8$\times$8 to 16$\times$16).}
    \label{fig:experiment3}
\end{figure}

Assessing the scalability of a robot swarm is essential, as real-world environments can span large spatial extents. In Experiment III, we therefore evaluate the performance of our algorithm across arenas of varying sizes. As illustrated in Fig.~\ref{fig:experiment3}, all methods exhibit increasing completion times as the arena grows, reflecting the fundamental scaling challenge of distributed foraging: larger workspaces reduce encounter rates (lower robot density per unit area) and increase travel distances between the nest, search trajectories, and discovered resources.

\paragraph{Small vs. Large Arenas:} In the smallest arena (8~$\times$~8), the robot density relative to the environment size is high, making resource discovery comparatively easy and limiting the opportunity for any exploration policy to differentiate itself. As arena size increases, performance degrades for all three methods because robots must (i) spend more time in transit, (ii) cover a larger area before first contact with resources, and (iii) manage a growing fraction of the environment that has been visited but is not explicitly remembered.

CPFA degrades most sharply because its exploration relies heavily on repeated uninformed searches and local, short-horizon memory (site fidelity and pheromone attraction), which can lead to redundant revisits and delayed coverage of distant, unvisited regions when the workspace expands. GPFA mitigates some redundancy by suppressing revisits within a coarse 3$\times$3 grid, but this fixed discretization does not adapt to arena scale: as the arena grows, a larger number of grid cells must be touched before coverage is achieved, and the method can still allocate effort to uniformly exploring low-yield space.

In contrast, ARPC maintains a more consistent scaling trend by explicitly representing already-explored space via clustered repulsive pheromones and biasing informed search toward the complement (never-visited) regions. This mechanism reduces redundant exploration as the arena expands and helps the swarm allocate limited search time to frontier areas rather than repeatedly re-sweeping the interior. Quantitatively, ARPC requires on average an additional 404 seconds for each 2$\times$2\,m increase in arena size, compared to 519 seconds for GPFA, indicating improved scaling efficiency.

\vspace{-6mm}

\section{Conclusion}
\label{conclusion}

In this work, we introduced ARPC, a cooperative foraging algorithm that enhances swarm efficiency through the adaptive clustering of repulsive pheromones to prune previously explored regions. Evaluated against the standard CPFA and the recent GPFA baselines across diverse spatial distributions, resource densities, and arena sizes, ARPC demonstrated robust and statistically significant improvements. 

Our experimental results establish that ARPC consistently achieves lower cumulative completion times than both baselines, particularly as resource density and environment size increase. Critically, the performance gains are not limited to the endgame phase; ARPC accelerates initial discovery and sustains efficient exploitation throughout intermediate collection milestones. Furthermore, the algorithm scales favorably with arena size, successfully mitigating the performance degradation characteristic of larger search spaces. Ultimately, ARPC presents a highly effective strategy for large-scale, heterogeneous foraging, motivating future extensions toward real-world deployments involving dynamic targets, communication limits, and energy constraints.

\begin{credits}
\subsubsection{\ackname} The authors acknowledge the financial support provided by the NSF Expand AI program (No. 2434916), the NSF CREST Center for Multidisciplinary Research Excellence in Cyber-Physical Infrastructure Systems (MECIS) (No. 2112650), and the NSF MSI program (No. 2318682).

\end{credits}

\bibliography{references}
\bibliographystyle{splncs04}

\end{document}